\documentclass[10pt,a4paper]{article}
\usepackage[T1]{fontenc}
\usepackage[utf8]{inputenc}
\usepackage[margin=25mm]{geometry}
\usepackage{lmodern}
\usepackage{microtype}
\usepackage{booktabs}
\usepackage{amsmath}
\usepackage{graphicx}
\usepackage[hidelinks]{hyperref}
\usepackage[numbers,sort&compress]{natbib}
\newcommand{\paperauthors}{Yunus Serhat Bıçakçı}
\newcommand{\paperrepository}{\href{https://github.com/yunusserhat/firerisk}{github.com/yunusserhat/firerisk}}
\hypersetup{pdftitle={SigLIP2 for aerial fire risk classification},pdfauthor={\paperauthors}}
\title{SigLIP2 for aerial fire risk classification}
\author{\paperauthors\,\href{https://orcid.org/0000-0002-7288-9959}{%
  \raisebox{0.55ex}{\includegraphics[height=3mm]{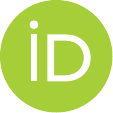}}}\\[0.4em]
  \small Department of Artificial Intelligence and Machine Learning\\
  \small Faculty of Applied Sciences, Marmara University\\
  \small Istanbul, Türkiye}
\date{}

\begin{document}
\maketitle

\begin{abstract}
We examine the transfer of a pretrained SigLIP2 image encoder to seven class
fire risk classification from aerial imagery. We introduce a reproducible
partition of the public FireRisk training mirror and an implementation that
records data provenance, preprocessing and model selection. Two initial runs
compare a frozen encoder probe with full model adaptation. On the validation
partition, full adaptation reaches 63.05\% accuracy and 58.94\% macro F1,
compared with 55.95\% and 50.19\% for the probe. Both runs use one training
seed and select their checkpoint on the same validation partition. These
development results support further evaluation of SigLIP2 but do not establish
performance on an independent test set or unseen regions. The accompanying
code provides a common framework for repeated experiments and comparisons
with additional visual encoders.
\end{abstract}

\section{Introduction}

Aerial images describe vegetation, buildings and surface cover that may help
distinguish wildfire hazard categories. FireRisk links such imagery to labels
derived from the Wildfire Hazard Potential map~\citep{shen2023firerisk}.
The source map represents the relative potential for wildfire that would be
difficult to suppress~\citep{dillon2020whp}. Recovering its categories from
imagery is an image classification task with a specific reference target.
It should be distinguished from forecasting ignition, seasonal fire occurrence
or the expected consequences of a fire.

The five ordered hazard levels make this problem more demanding than a
simple separation of water, vegetation and built surfaces. Similar vegetation
patterns can occur in tiles assigned different hazard levels. The labels
also come from a map that incorporates information beyond the displayed RGB
crop. A classifier may therefore recognize broad surface cues while missing
the distinctions that matter within the hazard categories. Aggregate accuracy
alone cannot show whether that distinction has been learned.

Large pretrained image encoders offer a practical starting point when a
task specific labeled corpus is limited. CLIP demonstrated transfer from
image and text supervision to varied visual recognition
tasks~\citep{radford2021clip}. SigLIP2 combines such supervision with
additional representation learning objectives~\citep{tschannen2025siglip2}.
This motivates a question about how its image encoder transfers to aerial
imagery. Reusing a frozen encoder tests the information already present in
the representation. Updating the encoder allows that representation to
change with the hazard labels, but also changes optimization demands.

We examine class errors alongside aggregate accuracy, using a frozen encoder
probe and a fully adapted encoder under two declared training recipes.
The comparison asks whether the selected representation and optimization
settings improve recognition within the hazard categories, rather than only
the more distinct surface classes.

We make the available data and selection procedure explicit because the
public mirror does not contain the original evaluation partition. The study
provides a reproducible split, audited data access, two initial adaptation
results and an analysis of their errors. Its purpose is to establish a
transparent starting point for further comparisons. It does not claim a
new best result for the published FireRisk benchmark or a validated mapping
system for unseen locations.

\section{Related work}

FireRisk provides supervised and self supervised representation benchmarks
for aerial fire hazard classification~\citep{shen2023firerisk}. We examine
transfer from a more recent pretrained image encoder under a separately
declared partition.

Pretraining tailored to Earth observation addresses a different source of
variation from a generic visual corpus. SatMAE incorporates temporal
embeddings and spectral position encodings when learning from satellite
observations~\citep{cong2022satmae}. Its temporal and multispectral inputs
contain information unavailable in the RGB FireRisk mirror. RemoteCLIP
adapts vision and language learning to remote sensing through image captions
derived from heterogeneous annotations and the inclusion of aerial
imagery~\citep{liu2024remoteclip}. These studies motivate attention to the
pretraining domain. Their published results do not establish performance
under our split, and neither model was evaluated in the two runs reported
here.

Learning from aerial images also has a substantial dense prediction setting.
Bıçakçı and Sarıca combined attention gates and a Transformer within a
U Net for building segmentation from aerial imagery and laser
data~\citep{bicakci2023attransunet}. Their work illustrates the use of local
and broader contextual representations in aerial scene interpretation.
The present experiment instead uses RGB imagery alone and assigns one
hazard category to each tile. Building masks and fire hazard labels are
different targets, so segmentation performance does not imply hazard
classification performance.

SigLIP trains image and text representations using a pairwise sigmoid
objective~\citep{zhai2023siglip}. SigLIP2 extends this family with captioning,
self distillation, masked prediction and data curation
objectives~\citep{tschannen2025siglip2}. These encoders offer representations
that can be reused beyond their original image and text training task.
In a geospatial application, Bıçakçı and colleagues used SigLIP image
embeddings to retrieve gallery geolocations for multimodal language model
prompts~\citep{bicakci2025geolocalization}. That study keeps the pretrained
encoder for street imagery retrieval. Here we examine supervised adaptation
of SigLIP2 to aerial hazard labels. We use only its image encoder and do
not evaluate text prompts, language generation or geographic retrieval.

A complementary comparison concerns visual features learned without text.
DINOv2 uses self supervision and curated images to learn general visual
representations~\citep{oquab2023dinov2}. The code includes a DINOv2 preset
under the same data protocol. That makes a later comparison feasible, but
does not provide a completed DINOv2 result in this paper. Comparing general
visual encoders, language supervised encoders and models trained specifically
for remote sensing would help distinguish the importance of pretraining
content from adaptation strategy.

\section{Data and protocol}

The original FireRisk study describes 91,872 images, with 70,331 training
examples and a separate validation partition~\citep{shen2023firerisk}. Its RGB
imagery was obtained from the National Agriculture Imagery Program
(NAIP)~\citep{usda2025naip} during 2019 and 2020 and cropped to
$320\times320$ pixels. Labels come from the 270 m grid of the
2020 Wildfire Hazard Potential product~\citep{dillon2020whp}. We use the
\href{https://huggingface.co/datasets/blanchon/FireRisk}{\texttt{blanchon/FireRisk}}
mirror maintained by Julien Blanchon~\citep{blanchon2023firerisk}, which
exposes only the 70,331 training examples. The mirror therefore
does not provide the published evaluation partition.

\begin{figure}[!htb]
\centering
\refstepcounter{figure}\label{fig:datasetexamples}
\includegraphics[width=\textwidth]{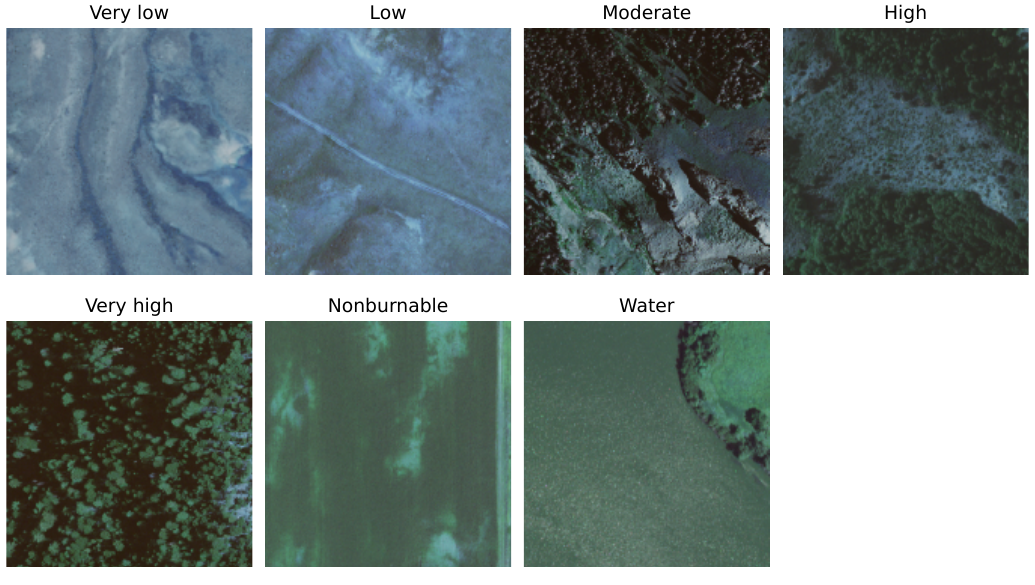}
\medskip
\parbox{\textwidth}{\small Figure~\thefigure. Deterministically sampled
training images with their provided FireRisk labels. Original imagery is
from USDA NAIP~\citep{usda2025naip}, compiled by Shen and
colleagues~\citep{shen2023firerisk} and accessed through the Blanchon
mirror~\citep{blanchon2023firerisk}. No test images were selected.}
\end{figure}

An audit of all 70,331 stored images confirms that each is a
$320\times320$ RGB PNG. Figure~\ref{fig:datasetexamples} shows one image per
class, selected uniformly from the training partition with seed 2026 and
no visual screening. These examples illustrate the inputs and assigned
labels. They do not establish that hazard can be inferred reliably by
visual inspection, or that one example represents an entire class.

We create a new split with 49,231 training images, 10,552 validation images
and 10,548 reserved test images. Allocation is stratified by class, using
split seed 2026. The seven categories are very low, low, moderate, high, very
high, nonburnable and water. The split preserves the class imbalance. For
example, the validation partition contains 3,264 very low examples and 490
very high examples.

\begin{figure}[!htb]
\centering
\refstepcounter{figure}\label{fig:datasetcounts}
\includegraphics[width=\textwidth]{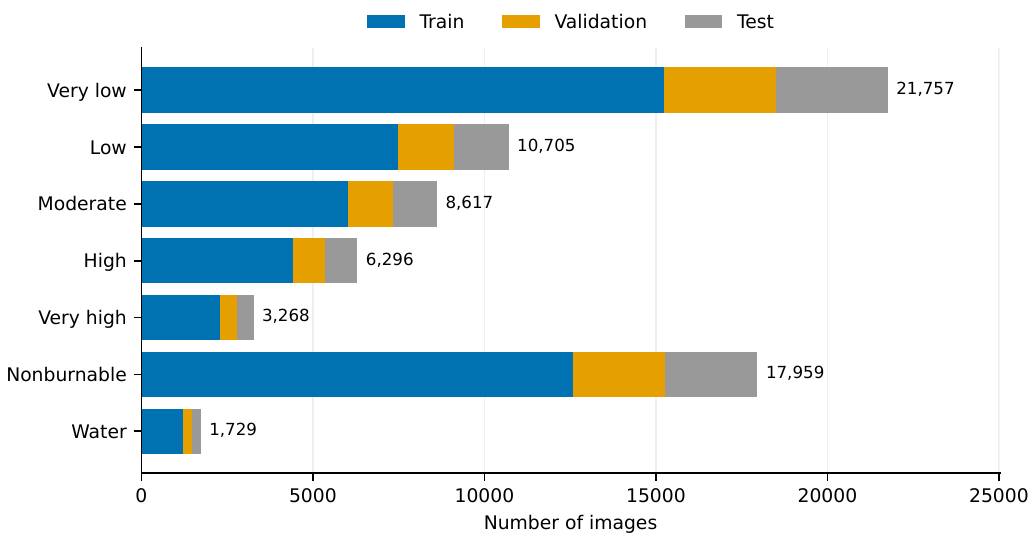}
\medskip
\parbox{\textwidth}{\small Figure~\thefigure. Class counts in the pinned
FireRisk mirror and their allocation to training, validation and reserved
test partitions. Counts describe the input corpus. No test predictions
were used to construct this figure.}
\end{figure}

Figure~\ref{fig:datasetcounts} reports the complete mirrored class counts
and their allocation to the new partitions. The split preserves the class
frequencies approximately, with rounding to whole images. Nonburnable and
very low together account for 56.47\% of the mirror, while very high
accounts for 4.65\%. This makes class level analysis central to evaluating
an aggregate improvement.

Very low accounts for 30.93\% of the validation partition, compared with
4.64\% for very high and 2.46\% for water. Predicting the largest class for
every image would therefore produce 30.93\% validation accuracy. We retain
the observed class frequencies instead of resampling the validation set.
Macro F1 gives each category equal weight, while class recall exposes which
categories are missed.

Before allocation, the preparation code hashes decoded RGB pixels and image
dimensions using SHA256 after correcting image orientation. It groups exact
duplicates and rejects conflicting labels within a group. The audit found
70,331 unique pixel groups and no repeated images in this revision. This
check concerns exact identity. It cannot exclude overlapping tiles or visual
similarity between nearby locations. The mirror contains image and label
fields without coordinates or acquisition dates, so this experiment cannot
establish geographic or temporal independence.

The data revision, model revision and split hash are recorded with each run.
Validation macro F1 selects the checkpoint. The reserved test partition has
not been evaluated in this study. As a result, our scores are development
measurements under a new protocol and cannot be compared directly with the
original FireRisk benchmark.

\section{Model and training}

\subsection{Classifier and objective}

We use the SigLIP2 base image encoder with $16\times16$ patches at
$224\times224$ input resolution, distributed through
\href{https://huggingface.co/timm/vit_base_patch16_siglip_224.v2_webli}{timm}.
The pretrained attention pooling is retained. A trainable LayerNorm and a
seven output linear layer operate on its 768 dimensional representation.
The resulting classifier has 92,891,143 parameters. In the frozen encoder
probe, only the head changes, leaving 6,919 trainable parameters. The encoder
remains in evaluation mode during training. Full adaptation updates all
parameters and uses head dropout with probability 0.1.

Let $f_\theta(x)$ denote the pooled image representation.
At evaluation, the class probabilities are
\begin{equation}
  p(y=c\mid x)=\operatorname{softmax}\!\left(W\operatorname{LN}(f_\theta(x))+b\right)_c.
\end{equation}
The frozen recipe updates $W$, $b$ and the LayerNorm parameters while keeping
$\theta$ fixed. Its head is therefore more flexible than a strictly linear
map of raw encoder features. Both recipes minimize smoothed categorical
cross entropy
\begin{equation}
  \mathcal{L}=-\frac{1}{B}\sum_{i=1}^{B}\sum_{c=1}^{K}q_{ic}\log p(y=c\mid x_i),
  \qquad q_{ic}=(1-\varepsilon)\mathbf{1}[y_i=c]+\frac{\varepsilon}{K},
\end{equation}
where $K=7$, $B$ is the batch size and $\varepsilon=0.05$. No image and text
pairing objective is used during adaptation.

\subsection{Preprocessing and optimization}

Training uses random crops covering 80\% to 100\% of the image area, with
aspect ratios from 0.9 to 1.1. Horizontal and vertical flips have probability
0.5, and rotations use multiples of 90 degrees. Evaluation uses deterministic
bicubic resizing and the pretrained center crop policy with crop proportion
0.9. Each RGB channel is normalized with mean 0.5 and standard deviation 0.5.

Both runs use AdamW with weight decay 0.05, label smoothing 0.05 and gradient
clipping at norm 1. Learning rates follow a cosine schedule after warmup over
10\% of the configured optimization steps. The probe uses a head learning
rate of $10^{-3}$ and at most 15 epochs. Full adaptation uses $10^{-5}$ for
the encoder and $3\times10^{-4}$ for the head, with at most 30 epochs.
No class weighting or mixup is applied. The effective batch size is 64 in
both cases. Full adaptation accumulates gradients over two batches of 32
images and uses gradient checkpointing. Computation uses bfloat16 on one
NVIDIA RTX 5090 GPU. Training seed 42 is shared by the two runs.

Early stopping follows seven epochs without a validation macro F1 improvement
greater than 0.0001. Both runs stop after 14 epochs and select epoch 7. The
different learning rates, dropout and schedule lengths make this a comparison
of two initial recipes. They prevent attribution of the observed difference
to encoder adaptation alone.

\section{Preliminary results}

Table~\ref{tab:results} reports uncalibrated predictions from the selected
checkpoints. Full adaptation improves validation accuracy by 7.10 percentage
points and macro F1 by 8.75 percentage points relative to the probe. Macro F1
assigns equal weight to the seven class F1 scores, making the minority classes
visible alongside accuracy.

\begin{table}[!htb]
\centering
\small
\refstepcounter{table}\label{tab:results}
\begin{tabular}{lrrrrr}
\toprule
Recipe & Trainable & Accuracy & Macro F1 & Time & Memory \\
       & parameters & (\%) & (\%) & (min) & (GiB) \\
\midrule
Frozen encoder probe & 6,919 & 55.95 & 50.19 & 12.1 & 0.74 \\
Full adaptation & 92,891,143 & 63.05 & 58.94 & 19.0 & 1.96 \\
\bottomrule
\end{tabular}
\medskip
\parbox{\textwidth}{Table~\thetable. Validation results from one training
seed. Time covers the training
loop, validation and checkpoint writing, excluding data preparation and
model loading. Memory is peak PyTorch allocated GPU memory and does not
represent total device or system memory.}
\end{table}

\subsection{Errors within hazard categories}

Table~\ref{tab:classes} shows the validation support, F1 and recall for each
class. Full adaptation has higher F1 for every category in these selected
checkpoints, but recall does not improve uniformly. High recall changes from
46.56\% to 45.08\% even though its F1 increases. This is a useful distinction
when missed hazard areas have different costs from false alarms.

\begin{table}[!htb]
\centering
\small
\refstepcounter{table}\label{tab:classes}
\begin{tabular}{lrrrrr}
\toprule
Class & Support & \multicolumn{2}{c}{F1 (\%)} & \multicolumn{2}{c}{Recall (\%)} \\
      &         & Frozen & Full & Frozen & Full \\
\midrule
Very low    & 3,264 & 60.86 & 69.87 & 59.56 & 68.96 \\
Low         & 1,606 & 35.35 & 41.76 & 36.80 & 44.83 \\
Moderate    & 1,293 & 29.40 & 42.03 & 25.91 & 42.00 \\
High        &   945 & 40.27 & 46.53 & 46.56 & 45.08 \\
Very high   &   490 & 21.29 & 39.90 & 14.49 & 33.47 \\
Nonburnable & 2,694 & 80.62 & 85.12 & 85.93 & 86.30 \\
Water       &   260 & 83.53 & 87.33 & 80.00 & 86.15 \\
\bottomrule
\end{tabular}
\medskip
\parbox{\textwidth}{Table~\thetable. Class results from the same 10,552 validation images. Support is
the number of true examples. Frozen denotes the frozen encoder probe and
Full denotes full adaptation.}
\end{table}

Full adaptation correctly identifies 164 of the 490 very high validation
examples. Its 33.47\% recall limits the usefulness of the current classifier
when missing high hazard areas has a substantial cost.

\begin{figure}[!htb]
\centering
\refstepcounter{figure}\label{fig:confusion}
\includegraphics[width=\textwidth]{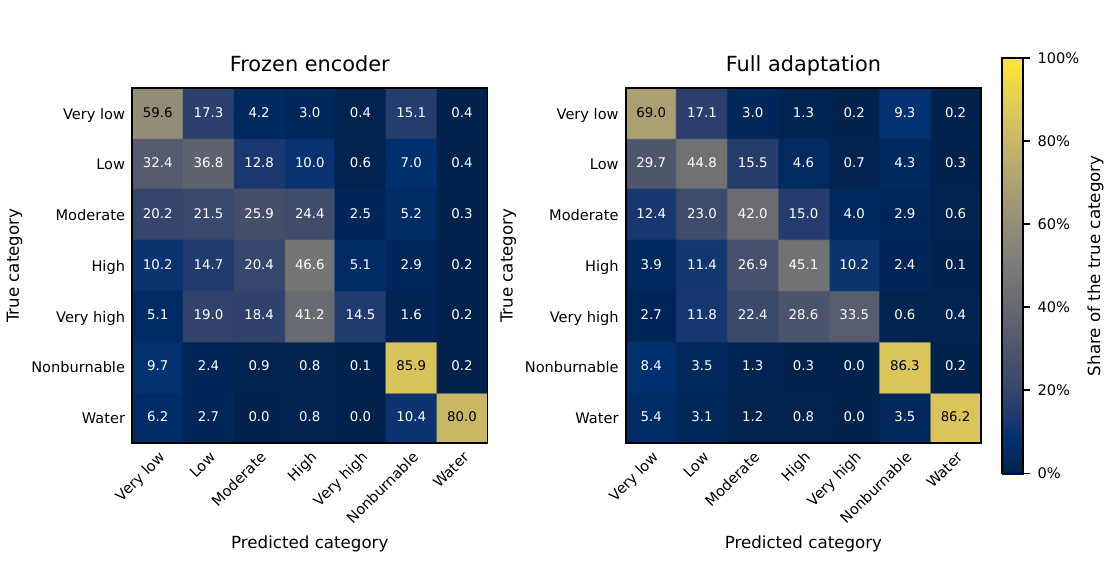}
\medskip
\parbox{\textwidth}{\small Figure~\thefigure. Validation confusion matrices normalized within each true class.
Rows show true labels and columns show predictions. Every prediction is
included, and each row sums to 100\% before rounding.}
\end{figure}

Figure~\ref{fig:confusion} shows that the remaining errors often fall into
another hazard category. Under full adaptation, 29.70\% of low examples are
assigned very low and 26.88\% of high examples are assigned moderate. Among
very high examples, 140 are predicted high and 110 moderate. Together these
two destinations account for 51.02\% of that class. Such errors can preserve
a broad vegetation or surface interpretation while losing the intended
hazard distinction. Establishing their physical cause would require image
review and additional covariates.

The 7,598 images whose true labels belong to the five ordered hazard levels
provide a more focused view. Exact category accuracy on these images rises
from 44.50\% to 54.01\%. Predictions of water or nonburnable are retained as
errors in this calculation. Of the 749 additional correct predictions
across the complete validation set, 723 come from these hazard classes.
The improvement is therefore not explained solely by easier water and
nonburnable recognition.

The runs executed concurrently on separate RTX 5090 GPUs. Their timings are
contextual measurements and do not provide a controlled speed comparison.
Times exclude environment installation, downloads and the initial pixel
audit. No repeated training seeds are available, so the variation caused by
training randomness remains unknown.

\section{Discussion}

The observed result is consistent with a useful pretrained representation
that benefits from task specific adaptation. It is also consistent with
differences in head optimization and schedule. Our two recipes do not
separate these explanations. A controlled follow up should match the search
budget and optimization schedule before estimating the contribution of
encoder updates. Independent seeds are needed to measure training variation.

The residual errors suggest a practical priority beyond increasing overall
accuracy. Very high recall remains 33.47\%, and confusion frequently moves
examples toward lower hazard categories. Class weighting, alternative
sampling or an ordinal objective are candidates for validation experiments.
They may trade precision against recall and should not be presented as
improvements without measured evidence. The current seven class loss treats
all incorrect labels categorically and does not impose a distance between
the five ordered levels. Water and nonburnable also require separate
treatment rather than placement on that ordinal scale.

Geographic independence remains unresolved. Ploton and colleagues showed
that ignoring spatial dependence can overstate predictive performance in
ecological mapping~\citep{ploton2020spatial}. That finding motivates caution
here, without establishing the size of any bias in this mirror. Pixel
deduplication cannot identify nearby or overlapping tiles. Coordinates and
scene identifiers would be needed for a spatial holdout, and acquisition
dates would support temporal evaluation. The model also learns map derived
hazard labels from RGB imagery. It has no weather, ignition or exposure
inputs, so the present evidence does not justify using its output as a
future fire probability or an operational warning.

\section{Reproducibility}

The implementation retains original Parquet shards and can index unchanged
image bytes for shuffled access without extracting individual files.
Each run saves its configuration, environment, preprocessing, data manifest,
history and selected weights. Immutable revisions and the split hash are
listed in the manuscript documentation. The initial artifacts record a
modified development tree but no source commit. Presets for LoRA, DINOv2,
ConvNeXt and ResNet50 are included without completed comparisons here.

The code separates source files from an explicitly chosen artifact directory
and provides a locked Python environment. Original shards occupy about
10.78 GiB, with a similar additional allocation for the optional image
archive. A shard access mode avoids this second allocation when disk space
is limited. The validation comparison can be regenerated from the saved
predictions. The dataset figures use the prepared manifest, split assignments
and training imagery. The supplied analysis scripts record their inputs and
numerical provenance.

Final experiments should use a committed release and declared validation
search budgets. The fixed recipes can then be evaluated on the reserved
test partition. Possible overlap with the pretraining data remains unknown.

\section{Conclusion}

In one preliminary comparison on a fixed FireRisk mirror split, full SigLIP2
adaptation reaches 58.94\% validation macro F1 against 50.19\% for a frozen
encoder probe. The gain extends to the five hazard categories, although
very high recall remains limited and high recall slightly decreases.
The protocol, code and class analysis provide a starting point for a
broader study. Repeated controlled experiments and independent evaluation
remain necessary before claiming reliable transfer to new regions.

\paragraph{Code and model availability}
Code is available at \paperrepository\ under the GNU General Public License
version 3. Initial
\href{https://huggingface.co/yunusserhat/firerisk-siglip2-base}{full adaptation}
and \href{https://huggingface.co/yunusserhat/firerisk-siglip2-base-frozen}{frozen encoder}
checkpoints on the Hugging Face Hub include inference instructions and
validation records. Upstream imagery and pretrained weights retain their terms.

\begingroup
\small
\raggedright
\setlength{\bibsep}{2pt}
\bibliographystyle{unsrtnat}
\bibliography{references}
\endgroup

\end{document}